(Review Article)

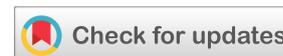

# Proactive fraud defense: Machine learning's evolving role in protecting against online fraud

Md Kamrul Hasan Chy *

*Department of Computer Information System & Analytics, University of Central Arkansas, Conway, Arkansas, USA.*

**Abstract**

As online fraud becomes more sophisticated and pervasive, traditional fraud detection methods are struggling to keep pace with the evolving tactics employed by fraudsters. This paper explores the transformative role of machine learning in addressing these challenges by offering more advanced, scalable, and adaptable solutions for fraud detection and prevention. By analyzing key models such as Random Forest, Neural Networks, and Gradient Boosting, this paper highlights the strengths of machine learning in processing vast datasets, identifying intricate fraud patterns, and providing real-time predictions that enable a proactive approach to fraud prevention. Unlike rule-based systems that react after fraud has occurred, machine learning models continuously learn from new data, adapting to emerging fraud schemes and reducing false positives, which ultimately minimizes financial losses. This research emphasizes the potential of machine learning to revolutionize fraud detection frameworks by making them more dynamic, efficient, and capable of handling the growing complexity of fraud across various industries. Future developments in machine learning, including deep learning and hybrid models, are expected to further enhance the predictive accuracy and applicability of these systems, ensuring that organizations remain resilient in the face of new and emerging fraud tactics.

## 1. Introduction

Online fraud can be described as any fraudulent activity that involves the use of the internet or digital technology [1]. Fraud has become a widespread issue across various sectors, with fraudsters continually finding new ways to exploit weaknesses in detection systems. Whether through identity theft, phishing, or other tactics, fraud with a digital or online element is getting harder to stop. While many fraud detection techniques have been developed and successfully used, fraudsters are quickly adapting and finding ways to outsmart these systems [2]. Traditional methods, such as rule-based detection or manual reviews, are now being exposed and outpaced by the increasingly sophisticated techniques used by criminals. This makes it clear that existing approaches, though effective in many cases, are struggling to keep up with the evolving nature of fraud. The gap in current research lies in the need for more adaptable and forward-thinking solutions that not only detect but also anticipate fraud across different types of cases. While some fraud detection systems have advanced to detect fraud as it happens, few have explored the potential of machine learning to fully prevent it before any damage is done. This research is important because it aims to address this gap by exploring how machine learning can improve fraud detection systems and offer a stronger defense by predicting fraud in real-time. This study offers a fresh approach that has not been fully explored, providing stronger protection for businesses and individuals against all forms of fraud.

* Corresponding author: Kamrul Hasan Chy

This paper explores how machine learning can enhance fraud detection systems, making them more effective and adaptable to modern threats. It focuses on how machine learning models, such as Random Forest and Logistic Regression, can improve the ability to identify complex fraud patterns and respond to evolving techniques used by fraudsters. By leveraging machine learning's ability to process large volumes of data and detect hidden patterns, this approach offers a more dynamic and proactive solution to fraud detection across various types of fraud, including identity theft and online scams. The research highlights how machine learning can be integrated into existing fraud detection frameworks to help businesses and organizations prevent fraud more efficiently. It also examines about machine learning's predictive capabilities can lead to stronger defense mechanisms, reducing financial losses and providing greater security [3].

Rather than merely reacting to fraudulent activities, the study highlights how machine learning algorithms can help businesses, organizations and customers to anticipate and prevent fraud before it occurs. This forward-thinking approach addresses the evolving nature of fraud and the limitations of existing methods, offering a more adaptable and effective solution. Another important contribution is the practical application of machine learning models to real-world fraud prevention strategies. The study outlines how these models can be integrated into existing frameworks, providing valuable insights for organizations seeking to strengthen their defense mechanisms against various forms of online fraud. This research not only advances the academic conversation but also provides actionable steps for improving security and reducing financial losses.

## 2. Literature Review

Fraud detection, particularly in the financial sector, has been a critical area of focus as fraudsters continuously find new ways to exploit systems. Supervised learning techniques have emerged as one of the foundational approaches in identifying fraudulent activities. These techniques involve training machine learning models on labeled datasets where transactions are categorized as either fraudulent or legitimate. Dornadula [4] explored the effectiveness of classifiers such as logistic regression and decision trees in detecting credit card fraud by learning patterns in historical transaction data. Their findings showed that supervised learning models could identify fraud with high accuracy, provided they are trained on substantial and well-labeled datasets. Similarly, Zareapoor [5] evaluated multiple supervised learning algorithms, including K-nearest neighbors (KNN) and Naive Bayes, demonstrating that KNN was particularly effective in detecting fraud in highly imbalanced datasets. Both studies underscore the crucial role of supervised learning in enhancing the accuracy of fraud detection systems, especially when dealing with large volumes of transaction data.

As the complexity of fraud schemes continues to evolve, hybrid models have become a critical innovation in fraud detection. These models combine the strengths of various machine learning algorithms to address the limitations of individual techniques, particularly in detecting fraud within imbalanced datasets. Pumsirirat [6] showcased the effectiveness of hybrid models by integrating autoencoders with restricted Boltzmann machines, allowing for better detection of fraudulent transactions in credit card data by learning from both labeled and unlabeled data. Similarly, Randhawa [7] explored the benefits of combining AdaBoost with majority voting methods to increase detection accuracy. Their research demonstrated that this ensemble approach significantly improved the performance of fraud detection systems, especially in scenarios with skewed datasets. These studies highlight how hybrid models offer a more flexible and reliable approach to detecting fraud, particularly in environments where traditional models may struggle.

In recent years, deep learning techniques have emerged as a powerful tool in fraud detection due to their ability to process vast amounts of data and identify complex patterns. These methods go beyond traditional machine learning by leveraging multi-layered neural networks to detect subtle anomalies that may be missed by simpler algorithms. Raghavan [8] explored the use of deep learning models, such as convolutional neural networks (CNNs), in detecting credit card fraud. Their study demonstrated that CNNs were highly effective in processing high-dimensional data and recognizing evolving fraud patterns. Melo-Acosta [9] also found success with deep learning approaches, particularly with autoencoders, which proved to be efficient in anomaly detection within financial datasets. Both studies underscore the advantages of deep learning models, particularly their capacity to handle large, complex datasets and adapt to new fraud techniques, making them a valuable asset in modern fraud detection systems.

With the increasing speed and complexity of financial transactions, real-time fraud detection has become a critical requirement for businesses aiming to prevent losses and protect consumers. This approach allows for the immediate identification and interruption of fraudulent transactions before they can cause damage. Dornadula [4] developed a real-time fraud detection system that integrates machine learning algorithms with live transaction data, enabling instantaneous detection and action. Their study highlighted how this system could significantly reduce financial losses by catching fraud at the point of transaction. Thennakoon [10] also emphasized the need for real-time detection, noting that predictive models capable of identifying fraud as it happens can help businesses prevent unauthorized transactions and minimize risks. Together, these studies demonstrate the growing importance of real-time systems in staying ahead of increasingly sophisticated fraud attempts and ensuring a faster response to potential threats.

Xuan [11] addressed the challenge of data imbalance in fraud detection, where fraudulent transactions are far fewer than legitimate ones, leading to skewed model performance. They tackled this issue by rebalancing the dataset to ensure that fraudulent transactions received adequate attention during analysis. Their aim was to improve the system's ability to detect fraud by overcoming the bias introduced by the overwhelming number of legitimate transactions. Similarly, Pumsirirat [6] implemented methods to enhance the detection of rare fraudulent transactions by adjusting how the data was processed. By focusing on strategies to balance the occurrence of fraud in the data, both studies demonstrated the importance of ensuring that fraud detection systems can accurately recognize less frequent, but critical, fraudulent activities. This focus on addressing data imbalance ensures that fraud detection efforts are not disproportionately concentrated on the more common, non-fraudulent cases, ultimately enhancing the overall effectiveness of fraud prevention systems.

Fraud detection systems must be dynamic to keep up with the constantly changing techniques used by criminals. Raghavan [8] highlighted the critical need for adaptive fraud detection models that can evolve alongside emerging fraud tactics. They pointed out that fraud patterns are not static and that detection systems must regularly adjust to remain effective. Similarly, Thennakoon [10] emphasized the value of continuously updating fraud detection systems to counter new and creative schemes used by fraudsters. Without regular updates, fraudsters may exploit the blind spots in outdated models, leaving businesses vulnerable. These studies collectively emphasize the importance of building detection systems that can quickly adjust to new patterns of fraud, ensuring that detection efforts remain relevant and effective in addressing modern threats.

## 3. Current State of Online Fraud

Online fraud involves the use of the internet and digital platforms to carry out deceptive activities targeting individuals, businesses, and financial institutions. Common forms of online fraud include identity theft, phishing, credit card fraud, and online scams. Identity theft, where personal information is stolen to impersonate someone else, leads to significant financial harm for both individuals and organizations. Phishing attacks, which trick people into providing sensitive information like passwords or credit card details, have also become widespread, especially with the increasing use of email and social media. Credit card fraud remains one of the most prevalent types of online fraud, allowing criminals to make unauthorized purchases with stolen card information. Additionally, online scams, such as fake e-commerce sites and fraudulent investment schemes, have led to direct financial losses for many unsuspecting victims.

The financial impacts of these fraudulent activities are severe. Ajayi [12] reports that global financial losses from cyber fraud amount to billions of dollars each year, affecting consumers and businesses alike. the overall economic impact of cybercrime, including online fraud, is projected to reach 10.5 trillion U.S. dollars by 2025 [13]. As fraudsters continue to refine their tactics, using tools such as social engineering, malware, and data breaches, they are increasingly able to avoid detection. Bolton, [2] notes that while traditional fraud detection methods have had some success, they are now struggling to keep pace with these evolving threats. This growing challenge underscores the need for more advanced and flexible fraud detection systems that can address the complexities of modern online fraud.

## 4. Traditional Fraud Detection Methods

Traditional fraud detection methods have long relied on rule-based systems, statistical analysis, and manual reviews to identify fraudulent activities. Rule-based systems flag transactions that deviate from predefined patterns, such as unusually large purchases or suspicious geographic locations, and statistical models detect anomalies by comparing new transactions to established norms. While these methods have been effective in many cases, they are limited in their ability to adapt to the constantly evolving tactics of fraudsters. Because these systems rely on fixed rules or historical data, they struggle to detect new types of fraud that do not fit into existing patterns. Baumann, [14] explains the scope of improvement in rule-based systems in detecting online fraud, highlighting their effectiveness but also their limitations. Manual reviews, which involve human auditors investigating flagged transactions, offer a more detailed analysis but are time-consuming, error-prone, and inefficient when faced with high transaction volumes. As fraud schemes become more sophisticated, traditional methods often fail to provide a proactive solution, detecting fraud only after it has occurred. Research by [15] underscores the limitations of these techniques, calling for more adaptable systems that can keep pace with the dynamic and growing threat of fraud.

## 5. Gaps in the Current Research

While machine learning has made significant strides in improving fraud detection systems, there are still notable gaps in the current research that need to be addressed. One major gap is the limited exploration of proactive fraud prevention. Most studies have focused on enhancing detection accuracy after fraudulent activities have occurred, rather than developing systems that can predict and stop fraud before it happens. Although machine learning models have shown potential in real-time fraud detection, the ability to fully prevent fraud through predictive analytics is still underexplored. Research by [2] highlights the importance of developing fraud detection systems that are not just reactive but also capable of anticipating fraudulent behavior before it leads to financial damage.

Another gap in the research is that most studies focus on specific types of fraud, like credit card fraud or phishing, without considering how machine learning models can be applied across various forms of fraud in different industries. While machine learning has been successful in detecting patterns in certain fraud types, there is a need for more generalized models that can handle a wider range of fraudulent activities. Additionally, there is a lack of collaboration between financial institutions and researchers, which limits the amount of data available for improving machine learning models [16]. Furthermore, there are very few studies that assess the long-term viability of these models across different financial services and types of fraud in the USA, which is crucial for understanding their effectiveness in real-world applications over time [16]. Addressing these gaps will be essential to developing fraud prevention systems that can adapt to the ever-changing landscape of fraud.

## 6. Proactive vs. Reactive Defense in Fraud Detection

Reactive defense mechanisms in fraud detection traditionally focus on identifying and responding to fraudulent activities after they have occurred. This approach utilizes rule-based systems and anomaly detection techniques that scan for deviations from predefined norms or patterns. Although reactive methods are essential for confirming and addressing breaches, they inherently lag behind the speed of evolving fraud tactics. Such systems often result in higher false positive rates and may struggle to adapt quickly to novel or sophisticated fraud schemes. However, incorporating advanced data visualization tools can enhance the effectiveness of reactive methods. As Chy & Buadi [17] note, visual tools can transform complex financial data into easily interpretable formats, enabling quicker decision-making and more effective identification of fraudulent activities, thereby mitigating potential losses more effectively.

In contrast, proactive defense strategies employ predictive analytics and advanced machine learning models to anticipate fraudulent activities before they occur. These systems analyze vast volumes of transactional data in real-time to detect subtle anomalies that might suggest a potential fraud. Utilizing machine learning techniques, such as those described in a research by Shan & Seunghwan [18], proactive strategies not only adapt to new and emerging threats but also significantly reduce the incidence of false positives, a common challenge in traditional fraud detection systems. By forecasting potential threats and identifying risky behaviors ahead of time, proactive models offer a dynamic and continually adapting solution that can preempt fraud attempts. This forward-thinking approach allows organizations to not only defend against known threats but also to anticipate and neutralize new threats as they develop, ensuring a robust defense mechanism that keeps pace with the rapid evolution of fraud tactics.

## 7. Emergence of Machine Learning in Fraud Detection

In recent years, machine learning has emerged as a powerful tool for enhancing fraud detection systems. Unlike traditional methods that rely on predefined rules and historical data, machine learning algorithms can process large volumes of data, identify complex patterns, and adapt to new fraud tactics in real time. Models such as Random Forest, Neural Networks, and Support Vector Machines have demonstrated significant improvements in detecting fraudulent activities. Bello [3] highlights that these models are particularly effective in identifying anomalies and detecting suspicious transactions that may not fit traditional patterns, making them well-suited to uncovering new and evolving fraud schemes.

One of the key advantages of machine learning is its ability to operate in real-time and reduce false positives, which occur when legitimate transactions are wrongly flagged as fraud. Verma [19] found that machine learning models are better at distinguishing between normal and abnormal behavior compared to traditional rule-based systems. Additionally, these models can learn and adapt as new fraud tactics emerge, offering a flexible and scalable approach to combating various types of fraud, from credit card fraud to more complex schemes such as identity theft and phishing. Studies by Bolton [2] and Bello [3] emphasize the transformative role that machine learning can play in making fraud detection more efficient and proactive.

## 8. Machine Learning Algorithms

Machine learning algorithms play a pivotal role in enhancing fraud detection systems by identifying patterns, anomalies, and trends in large datasets. These algorithms can be broadly categorized into supervised, unsupervised, and hybrid models, each with unique characteristics that make them effective for specific fraud detection tasks. Supervised models, like Logistic Regression and Decision Trees, are trained on labeled datasets where the outcome is known, enabling them to predict future fraudulent activities based on past patterns. Unsupervised models, such as K-means clustering, identify hidden structures and outliers in data without predefined labels, making them useful in detecting unknown fraud types. Hybrid models, which combine elements from both supervised and unsupervised techniques, aim to leverage the strengths of each approach, providing more robust and adaptive fraud detection systems. Each of these algorithms has its own strengths and weaknesses, making it essential to select the most appropriate model for a given scenario.

### 8.1. Logistic Regression

Logistic regression (LR) is a well-known and widely applied supervised machine learning algorithm, particularly suitable for binary classification problems like fraud detection. It models the relationship between one or more independent variables (such as transaction features) and a binary dependent variable (fraud or not fraud), outputting a probability score between 0 and 1. If the score exceeds a set threshold, the transaction is classified as fraudulent. LR's simplicity, ease of implementation, and interpretability make it a common choice for identifying fraudulent patterns in transactional data.

Several studies have demonstrated the effectiveness of logistic regression in fraud detection. Varmedja [20] highlighted that logistic regression, when combined with oversampling techniques like SMOTE (Synthetic Minority Over-sampling Technique), can improve its performance in handling imbalanced datasets, such as those seen in fraud detection where fraudulent transactions are rare. Additionally, Gupta [21] found that logistic regression was effective when applied to fraud detection in healthcare data, providing a benchmark model against which more complex algorithms were compared. The model's ability to handle large volumes of transactional data efficiently without requiring significant computational power has been one of its main strengths.

However, despite these advantages, logistic regression's linear nature means it may struggle to capture complex, nonlinear relationships in the data. For example, A study observed that while LR performed well on moderately imbalanced datasets, its accuracy decreased when handling more complicated, nonlinear fraud patterns, making it less effective in environments where fraudulent behaviors evolve rapidly [22]. As a result, logistic regression is often used in conjunction with more sophisticated algorithms or as a baseline model in fraud detection systems.

### 8.2. Decision Trees

Decision Trees (DT) are a popular choice for fraud detection due to their interpretability and ease of use. A decision tree works by recursively splitting data based on certain features, forming a tree-like structure where each internal node represents a decision rule, and each leaf node represents the final classification of a transaction as fraudulent or legitimate. Decision Trees are highly favored in fraud detection systems because they provide clear explanations for how decisions are made, making it easier for businesses to trust and understand the model's outcomes.

In a research, It was demonstrated that Decision Trees effectively classified fraudulent transactions by analyzing features such as transaction amount, location, and time. The model's simplicity allows it to handle both categorical and numerical data, making it adaptable to various fraud detection scenarios [5]. Additionally, Dornadula [4] pointed out that Decision Trees are particularly useful for classifying credit card transactions by examining user behavior and transaction history. However, one of the limitations of Decision Trees is their tendency to overfit the data, especially when applied to complex datasets. To address this, ensemble methods like Random Forest and Gradient Boosting are often used to improve model performance by combining multiple trees and reducing overfitting.

### 8.3. Random Forest

Random Forest is a powerful supervised machine learning algorithm widely used in fraud detection due to its ability to handle large and imbalanced datasets. It works by constructing multiple decision trees from random subsets of the data and combining their predictions to improve accuracy. This ensemble approach helps the model capture complex patterns and reduce the chances of overfitting, making it highly effective in identifying fraudulent activities. In healthcare fraud detection, for example, Random Forest has been applied successfully, achieving high accuracy and minimizing false positives when detecting fraudulent Medicare providers and claims [22,23].

One of the key advantages of Random Forest is its robustness in dealing with imbalanced datasets, a common issue in fraud detection where fraudulent cases are often a small fraction of the data. The algorithm's ability to handle this imbalance and provide accurate predictions has been demonstrated in various studies, including its use in universal healthcare schemes and large-scale financial fraud detection [4,21]. Additionally, Random Forest offers some level of interpretability, as individual decision trees within the forest can be analyzed to understand the model's decision-making process, making it a valuable tool in high-stakes areas such as healthcare fraud prevention [8].

**8.4. K-Nearest Neighbors (KNN)**

K-Nearest Neighbors (KNN) is a supervised learning algorithm that has proven effective in fraud detection, particularly due to its ability to classify instances based on proximity to known fraudulent activities. By measuring the distance between new transactions and past transactions, KNN assigns labels to new data points depending on their nearest neighbors. This method makes KNN particularly useful when fraudulent behaviors exhibit similar patterns to previously identified frauds. Studies have shown that KNN performs well when compared to other algorithms like Support Vector Machines (SVM) and Naive Bayes, particularly in environments where minimizing false alarms is essential [24].

One of the strengths of KNN in fraud detection lies in its simplicity and adaptability to imbalanced datasets. While dealing with high-dimensional data, KNN's performance can be enhanced through methods like feature selection and balancing techniques, such as Synthetic Minority Oversampling Technique (SMOTE). These techniques address the imbalance between fraudulent and non-fraudulent transactions, improving the algorithm's accuracy in detecting fraud. In one study, KNN demonstrated a high fraud detection rate when tested on a credit card dataset, effectively minimizing false positive rates compared to more complex models [25]. This adaptability, combined with its ability to detect anomalies in dynamic datasets, makes KNN a suitable candidate for real-time fraud detection systems.

**8.5. Support Vector Machine (SVM)**

Support Vector Machine (SVM) has become a popular choice in fraud detection due to its strong performance in identifying anomalies within complex and imbalanced datasets. Its ability to construct an optimal hyperplane that separates different classes allows SVM to handle diverse and non-linear data. This capability is especially useful in fraud detection, where fraudulent transactions often represent a small fraction of total data, making traditional classification approaches less effective. Recent studies have shown that SVM is highly efficient in fraud detection scenarios, such as detecting fraudulent Medicare claims, where the accuracy of identifying anomalies is crucial [23].

What sets SVM apart from other algorithms like Logistic Regression and Decision Trees is its flexibility in using kernel functions, which enables it to capture complex, non-linear relationships within data. By using these kernel methods, SVM can distinguish subtle differences in transaction behavior that may signal fraud. Additionally, research demonstrates that when combined with feature selection techniques, SVM not only improves accuracy but also enhances computational efficiency, making it an ideal solution for large datasets such as Medicare fraud detection [26]. This combination of high accuracy, low false positive rates, and scalability makes SVM a crucial tool in modern fraud detection systems. It ensures that fraudulent activities are detected with precision while minimizing unnecessary alerts, which improves efficiency in processing large volumes of data [27].

**8.6. Gradient Boosting**

Gradient Boosting is a robust machine learning algorithm known for its ability to build strong models by iteratively improving on the errors of weak learners, typically decision trees. This approach has been highly effective in fraud detection, particularly in the context of credit card fraud, where the data is often imbalanced, and fraud patterns are subtle. Taha [28] introduced an optimized LightGBM (a variant of Gradient Boosting), which outperformed traditional models like Random Forest and Logistic Regression. Their approach achieved an accuracy of 98.40% and a precision of 97.34%, making it highly suitable for real-world credit card fraud detection systems [28].

Furthermore, Gradient Boosting's flexibility in handling large datasets and evolving fraud patterns is one of its key strengths. The use of Bayesian optimization to fine-tune hyperparameters has been shown to enhance model performance significantly, especially in addressing the challenges of imbalanced datasets, as noted by Dal Pozzolo [29]. Gradient Boosting's ability to maintain high AUC scores—achieving 92.88% in Taha's [28] study—makes it a reliable method for minimizing false positives in fraud detection [29]. Additionally, other studies, such as those by [7], highlight Gradient Boosting's superior performance over models like AdaBoost and Majority Voting, further cementing its role as a leading technique in modern fraud detection [7].

**8.7. K-Means Clustering**

K-Means Clustering is a widely-used unsupervised machine learning algorithm, particularly effective in fraud detection due to its simplicity and scalability. The algorithm partitions datasets into k clusters based on similarities between data points. Each data point is assigned to the nearest centroid, which represents the mean of the cluster. In the context of financial fraud detection, K-Means is used to group transaction data into clusters that represent typical and atypical behaviors. Transactions that fall into outlier clusters, with characteristics differing significantly from the norm, are flagged for further investigation as potential fraud cases [30].

K-Means has been shown to excel in detecting patterns in large, high-dimensional datasets, a common challenge in the financial sector. For instance, in a study that applied K-Means to a dataset of over 500,000 transactions, the algorithm successfully identified outliers that represented fraudulent activities in less than 2% of the data. The ability to handle such imbalanced data is critical in fraud detection, as fraudulent transactions typically represent a small fraction of total transactions [30]. Additionally, clustering algorithms like K-Means offer significant computational efficiency, making them suitable for real-time fraud detection systems where processing speed is crucial.

One of the key strengths of K-Means is its flexibility. While the basic version of K-Means works well in many fraud detection scenarios, several variations have been developed to address its limitations, particularly in dealing with imbalanced datasets. Weighted K-Means and fuzzy K-Means are commonly employed to refine the clustering process by giving more importance to underrepresented classes, such as fraudulent transactions [31]. This improves the detection of fraud in datasets where the volume of non-fraudulent transactions dwarfs that of fraudulent ones, thus increasing the sensitivity of the model.

Despite its advantages, K-Means is not without challenges. The algorithm's performance can be highly sensitive to the choice of k (the number of clusters) and the initial placement of centroids. Methods such as the "elbow method" and the "gap statistic" are often used to determine the optimal number of clusters. In practice, choosing an incorrect k can lead to either over-clustering or under-clustering, impacting the model's accuracy in fraud detection. Nonetheless, when implemented correctly, K-Means is a highly effective tool in identifying fraudulent transactions, offering financial institutions a practical solution to mitigate the risks of financial fraud.

## 9. Machine Learning Model's Implementation

Using a Phishing Website Dataset containing 8955 entries and 31 features related to website characteristics, such as URL length, SSL certificate presence, and domain registration period, the Author conducted an analysis to develop predictive models for phishing website detection. These features were selected based on their relevance to identifying malicious websites, as phishing sites often exhibit certain patterns like long URLs or missing security certificates. The author's objective was to build machine learning models capable of accurately classifying websites as either legitimate or phishing, helping to mitigate the growing threat of online fraud. The dataset provided an extensive set of labeled data for training and testing.

The author's investigation employed a variety of machine learning models, with each model rigorously tested for accuracy. The results were as follows: Logistic Regression achieved an accuracy of 93.08%, Decision Tree Classifier scored 93.69%, and Random Forest Classifier stood out with a remarkable accuracy of 97.10%. Other models tested included K-Nearest Neighbors (KNN) with 63.99%, Support Vector Machine (SVM) at 57.17%, Gradient Boosting at 94.47%, AdaBoost Classifier with 91.74%, and Logistic Regression on Two Features at 85.71%. This variety of approaches allowed us to identify the best performers and those that might require further tuning or specific application conditions to enhance their performance.

The analysis revealed that ensemble methods like Random Forest and Gradient Boosting were particularly effective, due to their ability to manage the complex and imbalanced nature of the dataset. These models demonstrated a superior capability to discern subtle phishing indicators across the extensive feature set, underlined by their high accuracy scores. The Random Forest model, for instance, not only achieved the highest accuracy but also balanced precision and recall effectively, making it a robust choice for deploying in real-time phishing detection systems.

By applying hyperparameter tuning and optimization techniques, the author was able to further improve the performance of these models. Gradient Boosting, in particular, benefited from parameter adjustments that enhanced its ability to handle imbalanced datasets, where phishing websites represented a smaller proportion of the overall data. The predictive models developed through this analysis offer a scalable and highly accurate solution for real-time phishing website detection. By leveraging advanced machine learning techniques, these models can be integrated into online security systems to help reduce financial losses and enhance user protection against phishing attacks.

## 10. Future Scope of Machine Learning

The future scope of machine learning in fraud detection is vast, with the potential to revolutionize how organizations prevent and mitigate fraudulent activities. Machine learning algorithms, such as Random Forest and boosting techniques, are expected to further improve their ability to detect complex and evolving fraud patterns with greater accuracy and speed. As highlighted by Guo [32], these models can handle large, imbalanced datasets, continuously learning from new patterns to provide real-time fraud detection capabilities. Moreover, the integration of machine learning with organizational leadership strategies could further enhance its effectiveness. As diversity in top management teams has been shown to improve decision-making and investment efficiency Chowdhury [33], a diverse leadership approach could help guide the development and oversight of these machine learning systems, ensuring that different perspectives and innovative strategies are applied to strengthen fraud detection frameworks. This combination of technological advancement and diverse leadership involvement positions machine learning as a critical tool for safeguarding against the increasingly sophisticated nature of fraud in the future.

## 11. Conclusion

The growing complexity of online fraud underscores the urgent need for advanced, adaptive approaches to fraud detection. This paper has demonstrated how machine learning models, such as Random Forest, Decision Tree, Logistic Regression, Gradient Boosting etc. are transforming fraud detection by identifying intricate patterns in vast datasets and delivering real-time solutions [34]. Unlike traditional rule-based systems that react only after fraud has occurred, machine learning provides a proactive framework, enabling the detection and prevention of fraudulent activities before significant damage is done. By continually learning from new data, these models offer a highly efficient and scalable solution to combating fraud, making them invaluable in safeguarding financial systems and reducing the risk of financial losses.

As machine learning continues to evolve, its potential in fraud detection becomes increasingly evident. The ability of these algorithms to process and analyze data in real-time makes them a critical asset in environments where quick, accurate decisions are necessary to prevent fraud. Additionally, the flexibility of machine learning allows it to adapt to emerging fraud tactics, ensuring that organizations stay ahead of fraudsters. With a focus on improving accuracy and minimizing false positives, machine learning models offer businesses a more robust and dependable system for detecting fraudulent activities across various sectors, from financial transactions to online marketplaces.

Looking to the future, the scope of machine learning in fraud detection is vast, but continuous research and refinement are essential to fully harness its capabilities. Future developments should aim to create even more versatile and generalized models that can detect a wider range of fraud scenarios across multiple industries. Furthermore, enhancing the predictive power of these models will be crucial as fraud schemes become increasingly sophisticated. Machine learning's adaptability and scalability position it as a key tool in the ongoing battle against fraud, ensuring that organizations can better protect themselves and their customers from the ever-evolving threat of financial crime in the digital age.

## Compliance with ethical standards

*Disclosure of conflict of interest*

I declare that there are no conflicts of interest regarding the publication of this manuscript.